\documentclass[letterpaper, 10 pt, conference]{ieeeconf}  % Comment this line out if you need a4paper

\IEEEoverridecommandlockouts                              % This command is only needed if 
\usepackage{cite}
\usepackage{amsmath,amssymb,amsfonts}
\usepackage{graphicx}
\usepackage{algorithm}
\usepackage{algpseudocode}

\algrenewcommand\algorithmicindent{1.0em}%

\title{\Large \bf
STREAM: Software Tool for Routing Efficiently Advanced Macrofluidics
}

\author{Lehong Wang$^{1,2}$, Savita V. Kendre$^{1}$, Haotian Liu$^{1,2}$, and Markus P. Nemitz$^{1,3,4}$ % <-this % stops a space
\thanks{$^{1}$Robotic Materials Group, Department of Robotics Engineering, Worcester Polytechnic Institute, MA 01609, USA.
        {\tt\small \{lwang11,skendre,hliu8,mnemitz\}@wpi.edu}}%
\thanks{$^{2}$Department of Computer Science, Worcester Polytechnic Institute, MA 01609, USA.
        {\tt\small}}%
\thanks{$^{3}$Department of Electrical and Computer Engineering, Worcester Polytechnic Institute, MA 01609, USA.
        {\tt\small }}
\thanks{$^{4}$Department of Mechanical and Materials Engineering, Worcester Polytechnic Institute, MA 01609, USA.
        {\tt\small}}
}

\begin{document}

\maketitle
\thispagestyle{empty}
\pagestyle{empty}

\begin{abstract}

The current fabrication and assembly of fluidic circuits for soft robots relies heavily on manual processes; as the complexity of fluidic circuits increases, manual assembly becomes increasingly arduous, error-prone, and time-consuming.
We introduce a software tool that generates printable fluidic networks automatically.
We provide a library of fluidic logic elements that are easily 3D printed from thermoplastic polyurethanes using Fused Deposition Modeling only.
Our software tool and component library allow the development of arbitrary soft digital circuits. 
We demonstrate a variable frequency ring oscillator and a full adder. 
The simplicity of our approach using FDM printers only, democratizes fluidic circuit implementation beyond specialized laboratories. Our software is available on GitHub (https://github.com/roboticmaterialsgroup/FluidLogic)
% The software can be found on: "github.com/Lehong-Wang/Fluid-Circuit-Generator"

\end{abstract}

% \begin{IEEEkeywords}
% Keywords: 3D printing, soft robotics, fluidic computers, FDM printing, logic gate\end{IEEEkeywords}
Keywords: 3D printing, soft robotics, fluidic computers, FDM printing, logic gate

\section{INTRODUCTION}

With the emergence of interdisciplinary research, employing concepts from materials science, mechanical engineering, computer science, and biology, roboticists focus on creating robots made from soft, compliant materials rather than traditional rigid materials like metal or hard plastics. 
The field of soft robotics complements conventional robotics paradigms by exploiting the inherent advantages of soft materials, such as adaptability\cite{SoftGripper}, resilience\cite{soft_ring_oscillator}, resistance to damage\cite{GrowingTubeBot}, and simplicity\cite{SlitRobot}.

Currently, the majority of soft robots consist of soft structures for actuation, while the control system still relies heavily on rigid electronic components\cite{soft_robot_design_whiteside}. 
Since the predominant actuation mechanism for most soft robots is pneumatic-based\cite{penumatic_acuator}, there has been an increased research effort towards micro and macro fluidic circuits for facilitating new control mechanisms for soft robots\cite{FullPrintTurtle}. 

Techniques like E-beam lithography\cite{e-beam_review}, photo-lithography\cite{photolithography_micro_fluidic}, Ultraviolet (UV) exposure \cite{UV_micro_fluidic}, and PolyJet-based 3D printing\cite{PolyJet_Microfluidic} have been used for the fabrication of micro-fluidic circuits\cite{microfluidic_fabrication_review}.
They typically require specialized laboratories and prohibitively expensive supplies, limiting their impact in the field of robotics.
With micro-fluidic channels, the flow rate is limited to small flow rates (\(10^{-3}\) liters per minute), standing in sharp contrast to the high flow rates required by many soft robots\cite{soft_robot_flow_rate}.

\begin{figure}
\centerline{\includegraphics
[width=0.5\textwidth]{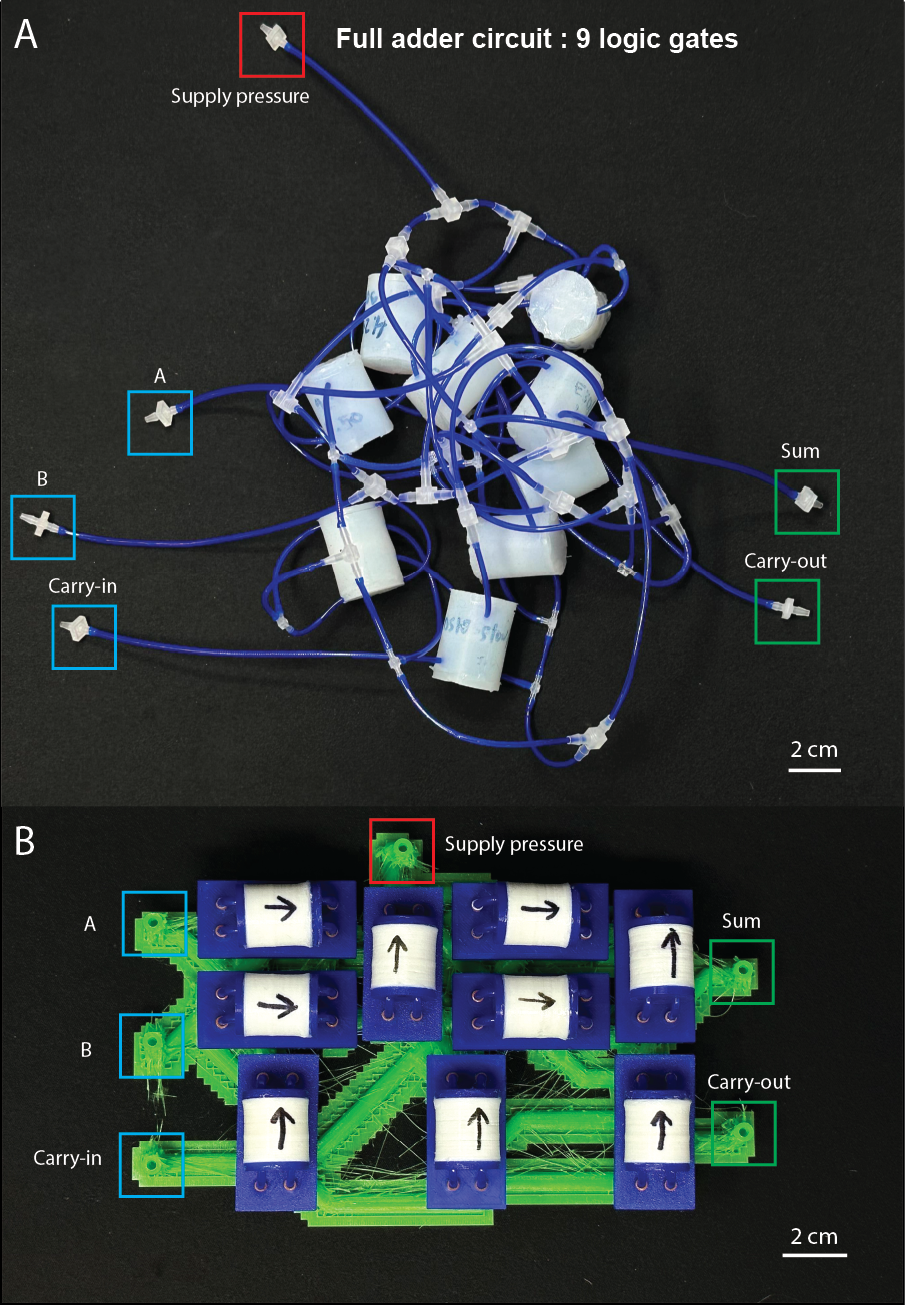}}
\caption{Two implementations of the same circuit design (full adder) with the traditional manual wiring approach (A) and our software enabled approach (B), which illustrates the impact of our software.}
\label{fig: problem_statement}
\end{figure}

Numerous research articles have explored the application of macro-fluidic circuits in the control of soft robots, primarily due to their ability to handle higher flow rates \cite{FluidicTurtle}.
One example is the introduction of the soft bistable valve\cite{BistableValControl}, which exhibits behaviors analogous to complementary metal-oxide-semiconductor (CMOS) devices and demonstrates the ability to control soft robots. 
Soft bistable valves are fabricated via replica molding; this fabrication process includes the creation of molds and the subsequent casting of materials to achieve the desired structural form, requiring technical expertise and specialized training for high quality results \cite{BistableValControl}.

Alternatively, low-cost logic elements have been created from tubes, straws, and balloons for less than \textcolonmonetary $50$ per device, but these components require manual fabrication processes as well \cite{Tracz2022Tube-BalloonElements}. Advanced tube-balloon type logic gates with CMOS-type switching behaviors increase the complexity of fabrication even further \cite{decker2022programmable}.

Recently, due to current advancements in 3D printing techniques, researchers have explored the field of 3D printed robots along with printable control systems.
A turtle-like robot was 3D printed in a single pass along with fluidic control elements \cite{FullPrintTurtle}.
A soft tunable bistable valve showcased the possibility of 3D printing logic elements \cite{TunableValve}.
These 3D printed devices required expensive printers (\(\geq\)\$100,000), making them inaccessible to many researchers. 
Recently, Zhai et al. used a low-cost Raise3D E2 industrial fused deposition modeling (FDM) printer (\(\approx\) \$4000) to design and fabricate soft, airtight pneumatic logic devices \cite{MikeTolly_zhai_desktopfabrication}. While this work encourages low-cost automated fabrication, little research has been conducted on automating the design process of fluidic control circuitry. A recent exception is work from Kendre et al., in which the \textit{Soft Compiler} is introduced \cite{SoftCompiler}. The \textit{Soft Compiler} aids the design of fluidic control circuitry for soft robots via a truth table; manual intervention is however still needed for the physical assembly of the circuit, making the overall creation of fluidic control circuitry for soft robots laborious. \\

In this study, we address the challenge of manually fabricating and assembling fluidic circuits.
Taking inspiration from printed circuit board (PCB) design, where the traces between circuit components can be automatically routed, we devised a software solution and a 3D printable implementation of a soft bi-stable valve to simplify the fabrication and assembly workflow of fluidic circuits \cite{BistableValControl}.
Our strategy calls for 3D printed, monolithic, fluidic circuits, onto which fluidic logic gates need to be affixed only, minimizing the margin for error and effort from manual assembly.

\section{IMPLEMENTATION}

Our proposed solution consists of a graphical software design tool based on \textit{Blender} and a set of fluidic circuit components that are developed to fit the software design flow. Blender is a free and open-source 3D creation software suite; it supports the modeling and rendering of 3D animations.
% The details of our fluidic circuit component design can be found in another paper from our group, which is still under review.
% FDM Printing: a Fabrication Method for Fluidic Soft Circuits? RoboSoft 2024

\subsection{3D printed version of a soft bistable valve}
Our redesigned soft bi-stable valve (\textbf{Figure 2A}) is printable with a FDM printer \cite{BistableValControl}.
The design utilizes the compliance and bi-stability of a membrane structure to apply a force to tubing on either side of a membrane.
The valve's body and membrane (\textbf{Figure 2A}) are printed from a white thermoplastic polyurethane (TPU) named \textit{Filaflex}, with a shore hardness of \(82A\)\cite{fliaflex_shore}.
The caps of the bistable valve (\textbf{Figure 2A}) are printed from a blue TPU (\textit{Ninjaflex}), with a shore hardness of \(85A\)\cite{ninjaflex_shore}.
The tubes are printed from \textit{Ninjaflex} as well, using a new type of custom print nozzle \cite{EthanPatent}. The nozzle enables a 3D printer to produce monolithic tubes with \(1.5~mm\) inner diameter and \(2.5~mm\) outer diameter. Alternatively, traditional silicone tubing can be used instead.

\begin{figure}
\centerline{\includegraphics
[width=0.5\textwidth]{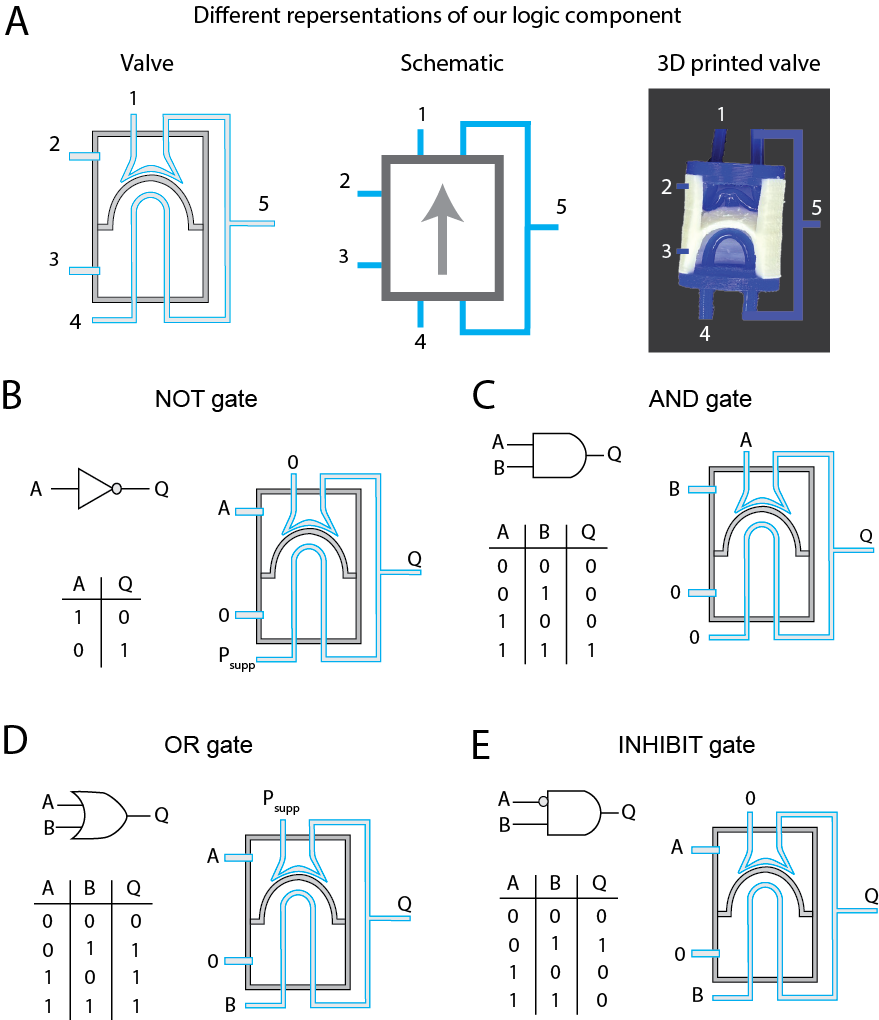}}
\caption{(A) three different representations of our 3D printable valve. They are used interchangeably in the article. (B-E) the different configurations of a 3D printable valve, including NOT, AND, OR, and INHIBIT logic gates. }
\label{fig:config_table}
\end{figure}

Our 3D printable version of a soft bistable valve mimics the working mechanism of a complementary metal-oxide semiconductor (CMOS) transistor pair \cite{CMOS}; the CMOS-based fluidic switch enables the design of fluidic circuits without the use of pull-down resistors\cite{pull_down}, circumventing power consumption at steady state and improving energy-efficiency. We re-engineered the valve design to transition from a labor-intensive (24 hours per device), multi-step fabrication process reliant on soft lithography, to a streamlined approach that requires only two hours for 3D printing and an additional 30 minutes for assembly.

\subsection{Tube routing algorithm}

The software is an add-on to the 3D modeling software Blender.
It uses Python to interface with Blender's built-in application programming interface (API).
To make the software design tool user-friendly, the user only needs to install Blender.
Our software package can be found on our GitHub repository.

The core of the software is a tube routing algorithm inspired by the \(A*\) path-finding algorithm: \(f(s) = g(s) + h(s)\)\cite{A_star}.
We implemented a customized \(A*\) algorithm.
The \(G\) and \(H\) functions are both given new path-finding weights.  The \(G\) function rewards downward or on-ground paths.
The \(H\) function enables the path-searching frontier to approach the destination by underestimating the distance to the destination.
To minimize the number of turns, the algorithm optimizes towards the direction with the largest gradient:

\begin{enumerate} 
    \item Identify states (in all dimensions \((x,y,z)\)) \begin{enumerate}
        \item \(s\): Current state
        \item \(s_p\): Parental state
        \item \(s_f\): Final/goal state
    \end{enumerate}
    \item Identify $G$ function: \(g(s) = g(s_p) + d(s_p, s)_{x,y,z} + R_1(s) + R_2(s)\) \begin{enumerate}
        \item \(g(s_p)\) is the \(G\) function to reach the parental state
        \item \(d(s_p, s)_{x,y,z}\) is the Euclidean distance from \(s_p\) to \(s\) for all dimensions
        \item Reward functions: \(R_1(s)\) is for downward, \(R_2(s)\) is for on-ground; \(dz = (s-s_p)_z\) represent the distance from \(s\) to \(s_p\) in \(z\) dimension
        
        \item For \(R_1(s)\): if \(dz < 0\) then \(R_1(s) = d(s, s_p)_{x,y} - d(s, s_p)_{x,y,z}\), which represents that \(R_1(s)\) is the difference between the Euclidean distance from \(s\) to \(s_p\) in \(x, y\) dimensions with the Euclidean distance from \(s\) to \(s_p\) in all dimensions; if in other \(dz\) conditions \(R_1(s) = 0\)

        \item For \(R_2(s)\): if \(z\equiv0\) then \(R_2(s) = -\alpha\) and \(\alpha\in(0,1)\) is a parameter to control reward value; if in other \(z\) conditions \(R_2(s) = 0\)
    \end{enumerate}
    \item Identify $H$ function: \(h(s) = m(s, s_f) + (\beta\cdot d(s, s_f)_{x,y,z}) \) \begin{enumerate}
        \item \(m(s, s_f) = max\{abs(s_f - s)_{x,y,z}\}\) represent the largest difference between \(s\) and \(s_f\)
        \item \(\beta\in (0, 1)\)
        \item \(d(s, s_f)_{x,y,z}\) represent the Euclidean distance from \(s\) to \(s_f\) for all dimensions
    \end{enumerate}
    \item Modified \(A*\) algorithm function: \(f(s) = (g(s_p) + d(s_p, s)_{x,y,z} + R_1(s) + R_2(s)) + (m(s, s_f) + (\beta \cdot d(s, s_f)_{x,y,z}))\)
\end{enumerate}

These changes enable the algorithm to run faster than the standard \(A*\) algorithm (evaluated by the number of explored nodes) while developing a tube routing path suitable for 3D printing.
\textbf{Figure 3} shows a comparison between the standard and our customized \(A*\) algorithm in generating path and searching areas; our algorithm is more efficient (searches fewer points) than the standard \(A*\) algorithm, producing a path that is designed for 3D printing.

\begin{figure}
\centerline{\includegraphics
[width=0.5\textwidth]{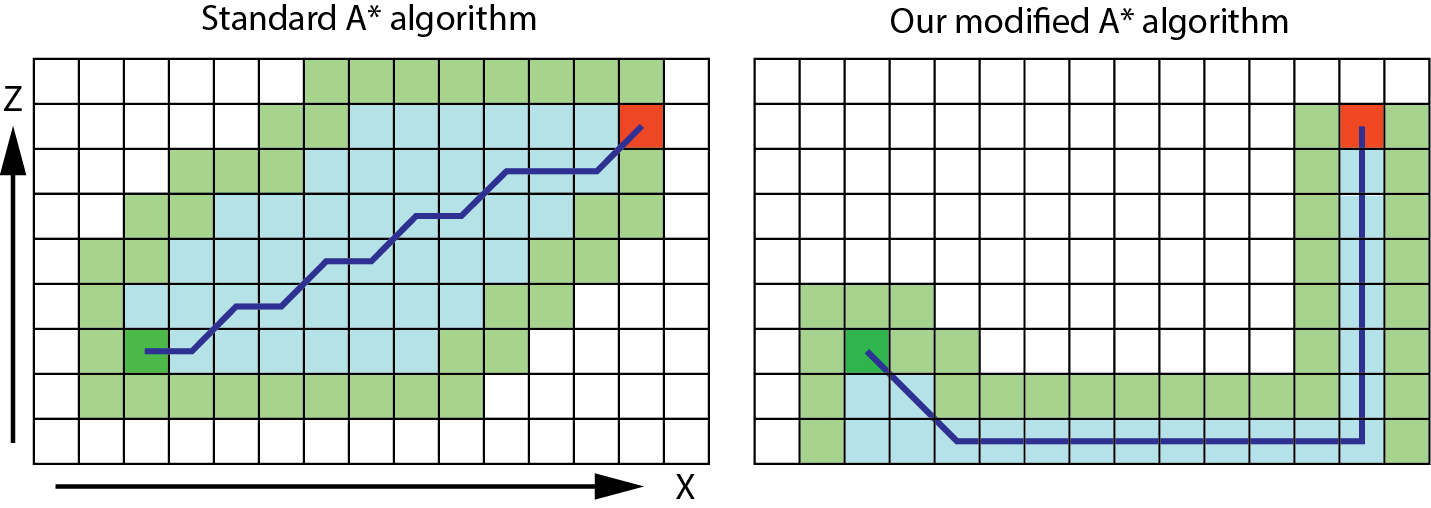}}
\caption{Comparing standard A* algorithm with our modified A* algorithm. Start node: deep green; End node: red; Explored nodes: blue; Frontier nodes: light green; Final path: deep blue line.}
\label{fig:algo}
\end{figure}

\subsection{Tube modeling algorithm}

\begin{algorithm}
\caption{Tube Network Model Construction}
\begin{algorithmic}

\For{$\forall (s, e) \in \mathcal{C}$} \Comment{$//\mathcal{C}$: set of connection pairs (start, end)}

    \For{$\forall p \in \{s, e\}$}
        \If{$ p \in \mathcal{P}$} \Comment{$//\mathcal{P}$: set of already connected points}
            \State $p \leftarrow p' \in Existing Path(p)$
            \State  \Comment{$//$Get a point from the existing path}
            \State Split existing path at $p'$
            \State $\mathcal{D}[p'] \leftarrow Junctioned Paths(p')$ 
            \State \Comment{$//\mathcal{D}$: dictionary of junction points and paths at $p'$}
        \EndIf
    \EndFor
    \State $path \leftarrow \text{Find Path }(s, e)$
    \State Record $path$
\EndFor

\For{$\forall path \in \mathcal{R}$} \Comment{$// \mathcal{R}$: set of recorded paths}
    \State $Make Tube Model(path)$
\EndFor

\For{$\forall j \in Keys(\mathcal{D})$}
    \State Add spherical connector $\mathcal{S}_j$ at $j$ to connect $\mathcal{D}[j]$
    
\EndFor

\State \Return Finished tube network model

\end{algorithmic}
\end{algorithm}

After finding the path for each set of connections, the path needs to be converted into 3D printable models.
Tubes are created from discrete lines, turned into tube surfaces with the Blender "\textit{bevel}" function, and finally turned into a mesh file with the Blender "\textit{solidify}" modifier. \textbf{Algorithm 1} explains its implementation in detail; the junctions between tubes are designed to enable the connection between three tubes from arbitrary directions by joining them in a hollow sphere.

\begin{figure}
\centerline{\includegraphics
[width=0.5\textwidth]{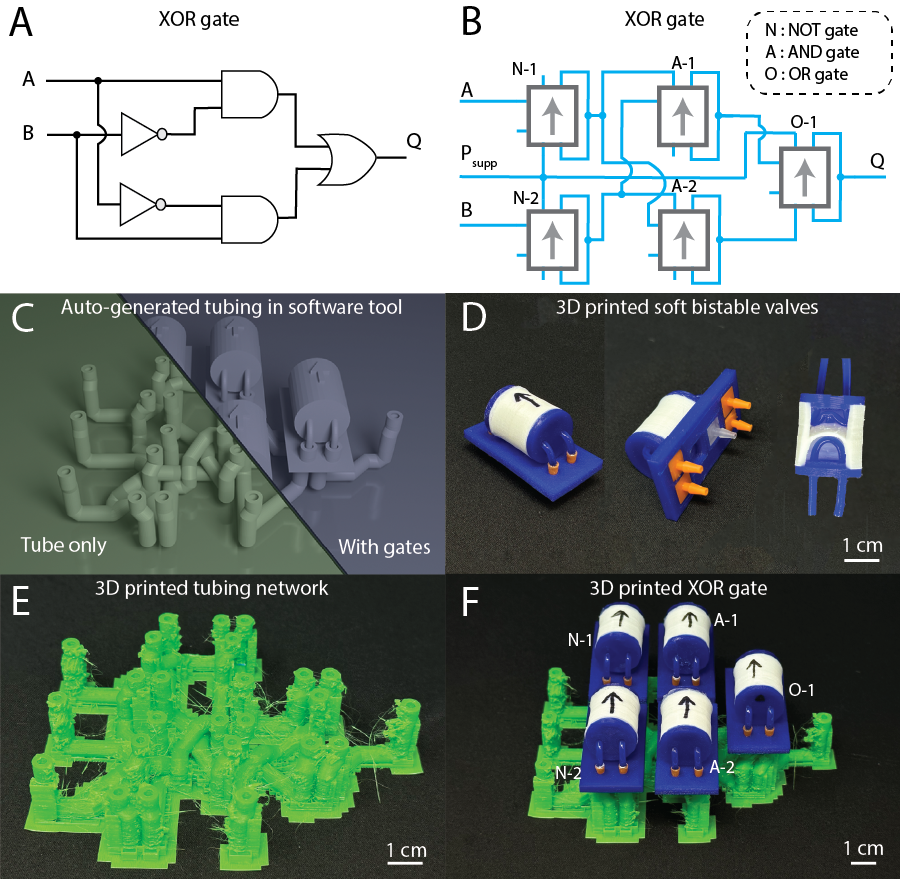}}
\caption{Process of fabricating fluidic circuit with our software. (A) logic diagram for the target circuit. (B) schematic for our printed valve implementation of the logic diagram. (C) rendered image of generated 3D models in our design software. The left half is shown without logic gates attached, the right half has logic gates attached (D) 3D printed valve component. (E) 3D printed fluidic network. (F) assembled fluidic circuit.}
\label{fig:process_chart}
\end{figure}

\subsection{Implementing circuit design in our Blender add-on}
In the software, the circuit elements can be found in our self-developed component library.
To add connections between components, the user selects circuit components and their port in the "\textit{Add Logic Gate Connection}" panel; there are also options for changing tube parameters. Renders of the 3D printable models of our variable frequency ring oscillator and full adder are shown in \textbf{Figures 4C, 5F}, and \textbf{6E}. Finally, the circuit design is exported to a ".stl" file, sliced, and 3D printed. To make the design workflow user-friendly, our Blender add-on implements a "script support" feature. By editing the script, users can change the circuit design model directly, avoiding the intermediate step of an user interface. A detailed tutorial can be found in our GitHub repository.

\subsection{Fabrication of tubing network}
To generate a circuit that can replace conventional tubes and tube connections, the circuit needs to satisfy the following conditions:

\begin{enumerate}

    \item The circuit tube tips must connect via a press fit.
    \item The fluidic circuit must be air-tight.
    \item The fluidic circuit must minimize fluidic resistance to allow for maximum air flow.
    \item The design of the fluidic circuit must be printable with a desktop FDM printer.

\end{enumerate} 

To satisfy these conditions, we conducted around five hundred tests of printed circuits and experimented with different thermoplastics and different print settings.
We concluded with a recommendation for a "Prusa MK3S" printer with \textit{Flexfill} (shore hardness 98A); the direct drive system and higher shore hardness filament allow for robust 3D prints of fluidic circuits satisfying our outlined conditions. Details on print parameters can be found in our GitHub repository.

\section{DEMONSTRATIONS}

We demonstrate the ability to fabricate combinational (memory-free) and sequential (memory-based) circuits. The circuits are designed as combinations of AND, OR, NOT, and INHIBIT logic gates. We provide videos of our circuit implementations in the Supplemental Information.

\subsection{Variable frequency ring oscillator}

A ring oscillator (F\textbf{igure 5}) is a device composed of an odd number of NOT gates (also known as inverters) in a ring whose output oscillates between true and false \cite{ring_osc}. The output of the last inverter is connected back to the input of the first inverter, creating a closed loop.
This leads to the continuous oscillation between true and false outputs in the ring, generating a clock signal (\textbf{Figures 5C and 5D}).
The frequency of the generated clock signal is determined by the combined propagation delay of the inverters and the time it takes for the signal to make one complete cycle around the ring \cite{ring_osc}.

\begin{figure}
\centerline{\includegraphics
[width=0.4\textwidth]{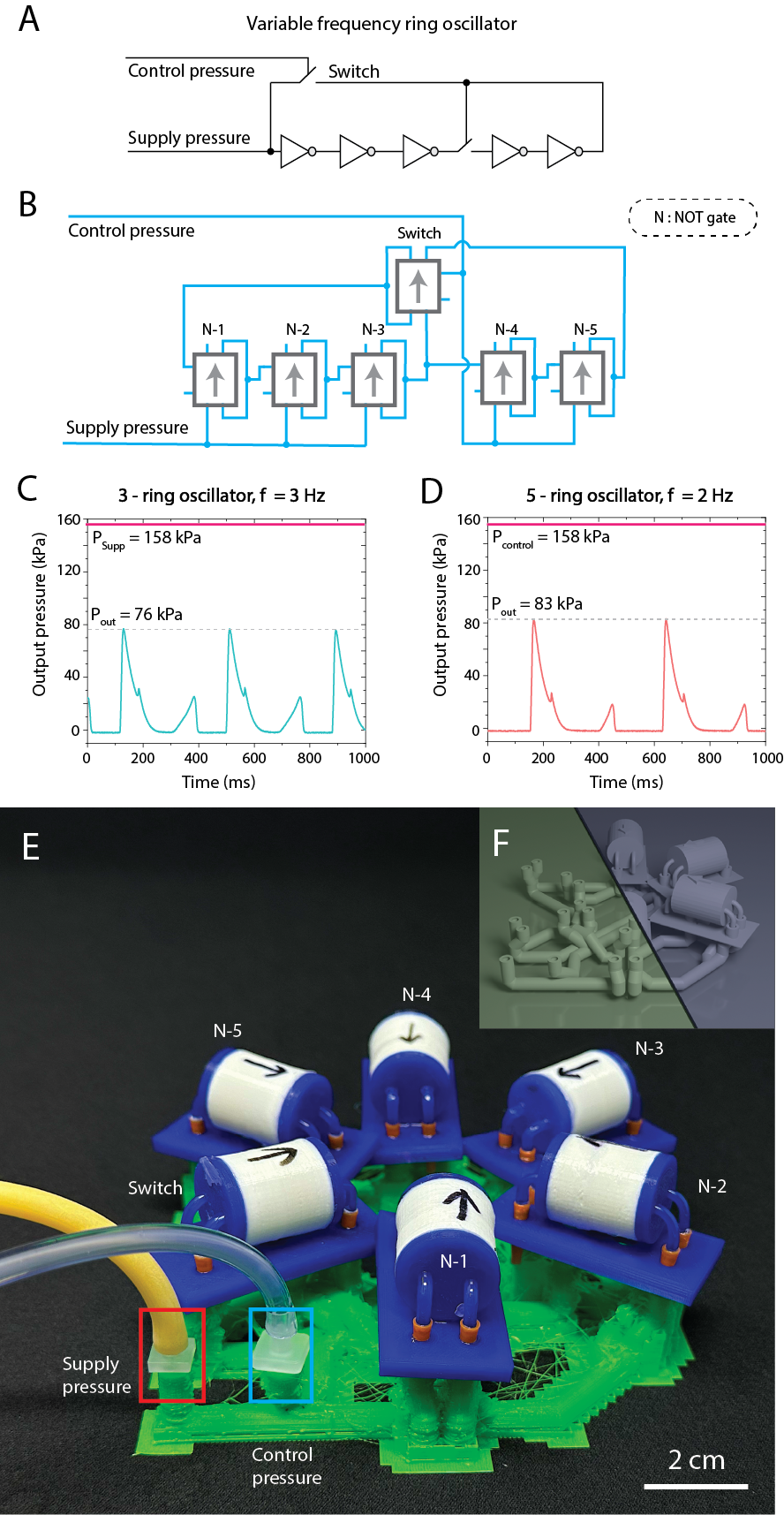}}
\caption{Variable frequency ring oscillator. (A) logic diagram (B) 3D printed valve schematic implementation of logic diagram. (C) output of ring oscillator with three gates and (D) five gates. (E) picture of the actrual experiment setup. (F) rendered picture of fluidic network in software.}
\label{fig:ring_oci}
\end{figure}

In this demonstration, we manipulate the control pressure to either an ON or OFF state, thereby altering the number of activated NOT gates within the loop. This adjustment enables us to generate signals of varying frequencies. Specifically, we engineered a pneumatic ring oscillator circuit comprising five NOT gates arranged in a loop. This loop incorporates a switch gate that dictates whether two of the five NOT gates will be bypassed during each cycle.

When the control pressure is in the OFF state, the loop connects only three NOT gates; when the control pressure is in the ON state, all five NOT gates are integrated into the loop. The signal of the output pressure peaks at \(76\,kPa\) in the three-ring configuration, and at \(83\,kPa\) in the five-ring configuration. The addition of two NOT gates results in a frequency shift of the output signal from $3~Hz$ to $5~Hz$.

\subsection{Full adder}

A full adder is a fundamental digital circuit that performs arithmetic addition on binary numbers\cite{mano_digital}.
It is a crucial building block in digital electronics and forms the basis for more complex arithmetic operations in computer processors and other digital systems.

A full adder consists of three main inputs ($A$, $B$, and $C_{in}$) and two outputs ($Sum$, $C_{out}$).
A and B represent the two summands.
$C_{in}$ (Carry-In) is the input that represents the carry bit from the previous stage when adding multiple binary numbers in cascaded full adders.
The sum is the output representing the least significant bit (LSB) of the sum of A, B, and $C_{in}$.
$C_{out}$ (Carry-Out) represents the carry bit generated when adding A, B, and $C_{in}$.

\begin{figure}
\centerline{\includegraphics
[width=0.45\textwidth]{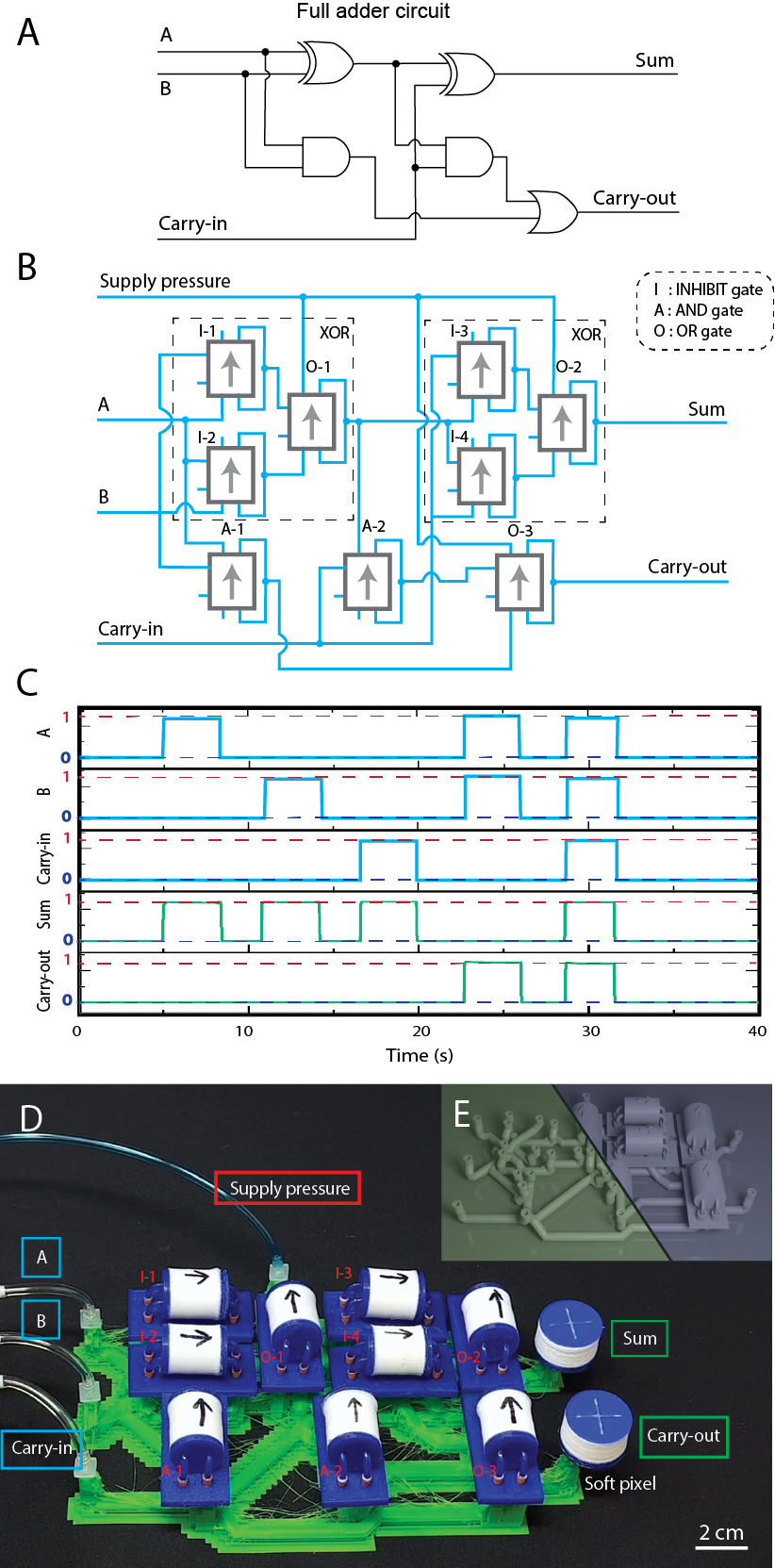}}
\caption{Full adder. (A) logic diagram (B) 3D printed valve schematic. (C) input and output pressure signals of the full adder. (D) picture of the actual experiment setup. (E) rendered picture of fluidic network in software.}
\label{fig: full_adder}
\end{figure}

The implementation of a full adder requires nine logic gates, fifty tubes, and sixty-six connectors \cite{fluidic_adder}.
This process is challenging to complete with manual assembly.
Using our approach, the entire assembly process only consists of attaching nine fluidic logic gates onto the 3D-printed fluidic circuit.
To visualize the pneumatic output of the full adder ($Sum$, $C_{out}$), we used a soft pixel, which will inflate and change color when inflated. Nemitz et al. developed the pneumatic display using soft lithography \cite{MarkusNemitz2020SoftMemory}; we converted the design into a 3D printed device.

\section{DISCUSSION}
\subsection{Towards an automated design and fabrication process of fluidic control circuits for soft robots}
This work depicts an important milestone towards the development of a completely automated design and fabrication process of fluidic controllers for soft robots. While this work addresses the issue of manually wiring pneumatic tubing; our previous work addressed the issue of creating fluidic circuit designs from scratch using the \textit{Soft Compiler} \cite{SoftCompiler}. We believe that future work includes systematic circuit analyses studying propagation delays, race conditions and the implementation of sequential logic, and limitations of single-layer circuits on a print bed and their stacking in three dimensions with anticipated increase of circuit complexity. 

\subsection{Additively manufacturing monolithic logic gates via FDM}
FDM is the most commonly used printing technique at a comparatively low system cost. Our work introduces a component library with soft bistable valves re-engineered for FDM printing. Currently, our valves can be 3D printed in two hours and assembled in half an hour. Future work should investigate the development of monolithic logic gates that yet possess CMOS characteristics without the use of expensive printers or complex post-processing techniques. Work on new types of nozzles combined with multi-material printing via tool changers is a promising research direction towards that goal. 

\subsection{What are fluidic circuits useful for?}
We demonstrate two fluidic circuits that were entirely 3D printed with low-cost printers. While fluidic circuits have been demonstrated for the control of soft robots, their limitations towards real-world applications have to be further explored. Although miniaturization or microfluidics will allow for some degree of control or intelligence, it will not compete with modern electronics. We believe the unique properties of fluidic circuits including their resistance to electromagnetic radiation and physical insults give them complementary features in future robotic systems. They may serve for redundancy at times when electronics has been jammed or destroyed, and mission completion can be satisfied with lower-tier intelligence.

\section{CONCLUSION}

This paper presents a software for the fabrication of pneumatic circuits and a 3D printable version of a soft bistable valve. A variation of the A* path planning algorithm  automatically generates 3D printable fluidic pathways between fluidic components. Once the circuit is printed, the user only needs to place components at pre-defined positions. We demonstrate a ring oscillator and a full adder as circuit implementations.  This work is another puzzle piece of a software stack that enables the automatic design and fabrication of fluidic control circuitry for (soft) robots. We share our code and component library via GitHub with the intend to create a community-driven research project.

\bibliography{main}
\bibliographystyle{IEEEtran}

\end{document}